\documentclass[pmlr]{jmlr}%

\usepackage{longtable}%

\usepackage{booktabs}
\usepackage[load-configurations=version-1]{siunitx} %

\makeatletter
\def\set@curr@file#1{\def\@curr@file{#1}} %
\makeatother

\theorembodyfont{\upshape}
\theoremheaderfont{\scshape}
\theorempostheader{:}
\theoremsep{\newline}

\jmlrvolume{126}
\jmlryear{2020}
\jmlrworkshop{Machine Learning for Healthcare}

\usepackage{amsmath}
\usepackage{dsfont}
\newcommand{\ind}{\mathds{1}}
\usepackage{float}
\usepackage{enumitem}

\title[Deep Kernel Survival Analysis and Survival Time Prediction Intervals]{Deep Kernel Survival Analysis and \\ Subject-Specific Survival Time Prediction Intervals}

\author{\Name{George H.~Chen}
       \Email{georgechen@cmu.edu}\\ 
       \addr Heinz College of Information Systems and Public Policy\\
       Carnegie Mellon University\\
       Pittsburgh, PA, USA}

\begin{document}

\maketitle

\begin{abstract}
Kernel survival analysis methods predict subject-specific survival curves and times using information about which training subjects are most similar to a test subject. These most similar training subjects could serve as forecast evidence. How similar any two subjects are is given by the kernel function. In this paper, we present the first neural network framework that learns which kernel functions to use in kernel survival analysis. We also show how to use kernel functions to construct prediction intervals of survival time estimates that are statistically valid for individuals similar to a test subject. These prediction intervals can use any kernel function, such as ones learned using our neural kernel learning framework or using random survival forests. Our experiments show that our neural kernel survival estimators are competitive with a variety of existing survival analysis methods, and that our prediction intervals can help compare different methods' uncertainties, even for estimators that do not use kernels. In particular, these prediction interval widths can be used as a new performance metric for survival analysis methods.
\end{abstract}

\section{Introduction}

Kernel survival analysis methods estimate subject-specific survival curves and times with the help of a kernel function, which measures how similar any two subjects are. Examples of such estimators include the conditional Kaplan-Meier estimator \citep{beran_1981}, random survival forests \citep{ishwaran_2008}, and survival support vector machines \citep{shivaswamy2007support,khan2008support}. When these estimators make a prediction for a test subject, they find the most similar training subjects and compute how much these training subjects contribute to the test subject's prediction. This information on the most similar training subjects could serve as a form of forecast evidence and could help in debugging.

How well a kernel survival analysis method works hinges on which kernel function is used. Phrased in a clinical context, defining how similar any two patients are is not straightforward and depends, for example, on what specific disease we are looking at and what the time-to-event outcome is (time until death, disease recurrence, hospital discharge, etc). To the best of our knowledge, the only existing methods for learning a kernel function for kernel survival analysis is to use a procedure like cross-validation to choose between pre-specified kernel functions (e.g., \citealt{cawley2004bayesian}), to automatically identify a weighted sum of pre-specified kernels \citep{dereli2019multitask}, or to use random survival forests \citep{ishwaran_2008}, which implicitly learns a kernel function in a greedy fashion (when growing trees) and has no known overall loss function that the method is minimizing.\footnote{For random forests (including its survival variant), the kernel function is, for any to feature vectors $x$ and $x'$, the fraction of trees for which $x$ and $x'$ are in the same leaf node \citep{breiman2000some}.}

In this paper, we present the first neural net framework that learns kernel functions for use with Beran's conditional Kaplan-Meier estimator (Section~\ref{sec:dksa}). Our approach adapts the neural kernel learning approach for classification by \citet{card2019deep} to the survival analysis setting. As with other neural survival analysis methods (e.g., \textsc{deepsurv} by \citet{katzman2018deepsurv}, \textsc{deephit} by \citet{lee2018deephit}), our approach requires a base neural net to be specified. We consider several choices that result in different neural kernel survival estimators of varying network depth, and we also discuss how to warm-start learning using either other neural survival estimators or random survival forests.

As a second contribution, we show how to construct prediction intervals for subject-specific survival time estimates (Section~\ref{sec:survival-conformal-prediction}). To do this, we use split conformal prediction \citep{papadopoulos2002inductive,lei2015conformal} and its weighted variant \citep{tibshirani2019conformal}. The former leads to prediction intervals that are valid \textit{marginally} (averaged over a whole test population) whereas the latter leads to prediction intervals that are valid \textit{locally} (averaged over subjects who are similar to a test subject according to a kernel function, such as one learned using our kernel learning framework or random survival forests). These intervals require a user-specified target coverage level $1-\alpha$ for $\alpha\in(0,1)$ (similar to confidence intervals). In a clinical context, prediction intervals that are locally valid with respect to a test subject are often more valuable than ones that only hold marginally: when a doctor tells a patient that the patient has a 90\% chance of recovery, we would like that 90\% to be averaged across individuals with attributes similar to the patient rather than across all individuals who might see the doctor.

In our numerical experiments (Section~\ref{sec:experiments}), we find that (a) our deep kernel survival estimators can achieve competitive prediction accuracy compared to existing survival analysis methods without taking longer to run, (b) our marginal and local subject-specific survival time prediction intervals have empirical coverage probabilities that closely match user-specified target coverage levels, and (c) we can use the width of our prediction intervals to compare different methods' uncertainties (the marginal prediction intervals can be used even for methods that do not use kernels) and to identify which subjects we are more uncertain about (for survival time estimators that do use kernels).

\subsection*{Generalizable Insights about Machine Learning in the Context of Healthcare}

Recent survival analysis advances in the machine learning community have focused on prediction accuracy, largely without worrying about interpretability of the learned models or how accurate predictions are at the subject-specific level. %
This paper makes progress toward resolving these two shortcomings. First, we combine some of the recent machine learning developments with kernel survival analysis, which is arguably more interpretable as it makes predictions based on finding which training subjects are most similar to a test subject. Second, we construct subject-specific prediction intervals that have statistical guarantees. %
We demonstrate our proposed methods on several standard publicly available healthcare survival analysis datasets that are on predicting time until death for various~diseases. %

\section{Background}

We begin by stating the standard survival analysis problem setup in Section~\ref{sec:survival-analysis-problem-setup} including providing notation and terminology used throughout the paper. We then review the conditional Kaplan-Meier estimator \citep{beran_1981} in Section~\ref{sec:conditional-kaplan-meier}, and split conformal prediction for constructing regression prediction intervals \citep{papadopoulos2002inductive,lei2015conformal,tibshirani2019conformal} in Section~\ref{sec:conformal-prediction}.

\subsection{Survival Analysis Problem Setup}
\label{sec:survival-analysis-problem-setup}

For ease of exposition, we phrase terminology using time until death as the outcome of interest; of course, other time-to-event outcomes can be used. We suppose we have access to $n$ i.i.d.~training subjects' data $(X_1,Y_1,\delta_1),(X_2,Y_2,\delta_2),\dots,(X_n,Y_n,\delta_n)$, where the $i$-th subject has feature vector $X_i\in\mathbb{R}^d$, nonnegative observed time $Y_i\ge 0$, and event indicator $\delta_i\in\{0,1\}$; $\delta_i=1$ means that the $i$-th subject's observed time $Y_i$ is a time of death, whereas $\delta_i=0$ means that the death time is missing and we only know that the $i$-th subject's time of death is at least $Y_i$ (the subject was still alive when data collection stopped). We assume there to be a distribution of feature vectors $\mathbb{P}_X$, a distribution of nonnegative survival times given a feature vector $\mathbb{P}_{T|X}$, and a distribution of nonnegative censoring times given a feature vector $\mathbb{P}_{C|X}$; these distributions are unknown. Each data point is assumed to be generated as follows:
\begin{enumerate}[leftmargin=1.5em,itemsep=-0.25ex,topsep=1ex] %
\item Sample feature vector $X_i\sim\mathbb{P}_X$.
\item Sample nonnegative survival time $T_i\sim\mathbb{P}_{T|X=X_i}$.
\item Sample nonnegative censoring time $C_i\sim\mathbb{P}_{C|X=X_i}$.
\item If $T_i \le C_i$ (death happens before censoring), set $\delta_i=1$ and $Y_i=T_i$; otherwise, set $\delta_i=0$ and $Y_i=C_i$. (In other words, $\delta_i = \ind\{T_i\le C_i\}$ and $Y_i=\min\{T_i, C_i\}$.)
\end{enumerate}
Using the training data, our goal is to estimate the conditional survival function $S(t|x):=\mathbb{P}(T>t|X=x)$ for any feature vector $x\in\mathbb{R}^d$ and time $t\ge0$; the function $S(\cdot|x)$ is the monotonically decreasing survival curve specific to a subject with feature vector~$x$.

Once we have an estimate~$\widehat{S}(\cdot|x)$ of $S(\cdot|x)$, we can estimate the survival time $T$ given $X=x$. To do this, we follow \citet{reid1981estimating} and find the time~$t$ where $\widehat{S}(t|x)$ crosses 1/2, which is an estimate of the median survival time for feature vector $x$. Specifically, we use
\begin{equation}
\widehat{T}(x):=\frac{1}{2}\big[\inf\{t\ge0 : \widehat{S}(t|x)\ge1/2\} + \sup\{t\ge0 : \widehat{S}(t|x)\le1/2\}\big].
\label{eq:median-surv-time}
\end{equation}
We provide more intuition for this estimator along with some other ways to estimate subject-specific survival times in Appendix~\ref{sec:surv-time-estimation}.

\subsection{Conditional Kaplan-Meier Estimators}
\label{sec:conditional-kaplan-meier}

Our proposed neural kernel learning framework for survival analysis builds on the conditional Kaplan-Meier estimator \citep{beran_1981}. To explain how this estimator works, we first explain the classical Kaplan-Meier estimator that estimates the \textit{marginal} survival function $S_{\text{marg}}(t):=\mathbb{P}(T>t)$ \citep{kaplan_meier_1958}.

\paragraph{Kaplan-Meier estimator}
The Kaplan-Meier estimator does not use feature vectors $X_1,\dots,X_n$ and only uses their observed times $Y_1,\dots,Y_n$ and event indicators $\delta_1,\dots,\delta_n$. We denote the sorted unique observed times as $t_1<t_2<\cdots<t_m$, where $m$ is the number of unique observed times. For time index $\ell\in\{1,2,\dots,m\}$, let $d_\ell$ be the number of deaths that occur at time $t_\ell$, and let $n_\ell$ be the number of subjects at risk right before time $t_\ell$:
\begin{equation}
d_\ell = \sum_{i=1}^n \delta_i \ind\{Y_i = t_\ell\},
\qquad
n_\ell = \sum_{i=1}^n \ind\{Y_i \ge t_\ell\}.
\label{eq:kaplan-meier-death-at-risk-quantities}
\end{equation}
Then the Kaplan-Meier estimate for marginal survival function $S_{\text{marg}}(t)$ is given by
\begin{equation}
\widehat{S}_{\text{marg}}(t) := \prod_{\ell=1}^{m} \Big(1-\frac{d_\ell}{n_\ell}\Big)^{\ind\{t_\ell \le t\}}\quad\text{for }t\ge 0.
\label{eq:kaplan-meier}
\end{equation}
This estimator has a natural interpretation: we multiply empirical probabilities of surviving from time 0 to $t_1$, from time $t_1$ to $t_2$, and so forth up to the given time~$t$. Note that the Kaplan-Meier estimator is usually stated such that the times $t_1,\dots,t_m$ are the unique times \textit{in which death occurred}. In our exposition to follow, it will be convenient to allow for times in which deaths did not occur. This does not affect the estimator: if there is no death at time $t_\ell$, then $d_\ell=0$ so $(1-\frac{d_\ell}{n_\ell})=1$, i.e., the product in equation~\eqref{eq:kaplan-meier} stays the same.

\paragraph{Conditional Kaplan-Meier estimator} To account for feature vectors, \citet{beran_1981} weight the contribution of different training data in the Kaplan-Meier estimator. As an example of this, given a feature vector $x$, we can find all training data within a pre-specified distance~$\sigma$ of $x$, and restrict the Kaplan-Meier estimator calculation to only use these training data. More generally, we weight each training data $X_i$ based on how similar $X_i$ is to the test feature vector $x$ using a kernel function $K$, where the similarity score between feature vectors $x$ and $x'$ is $K(x,x')\in[0,\infty)$. The example of only using training data within distance $\sigma$ corresponds to using the ``box'' kernel $K(x,x')=\ind\{\|x-x'\|\le\sigma\}$.

Instead of keeping track of the number of deaths and number of subjects at risk at different death times as in equation~\eqref{eq:kaplan-meier-death-at-risk-quantities}, we now instead keep track of their weighted versions:
\begin{equation}
d_K(t|x) := \sum_{i=1}^n \delta_i K(x, X_i) \ind\{Y_i = t\},\quad\;
n_K(t|x) := \sum_{i=1}^n K(x, X_i) \ind\{Y_i \ge t\}.
\label{eq:kernel-kaplan-meier-death-at-risk-quantities}
\end{equation}
Generalizing equation~\eqref{eq:kaplan-meier}, Beran's conditional Kaplan-Meier estimator is given by
\begin{equation}
\widehat{S}_K(t|x) := \prod_{\ell=1}^m \Big(1-\frac{d_K(t_\ell|x)}{n_K(t_\ell|x)}\Big)^{\ind\{t_\ell \le t\}}\quad\text{for }t\ge 0,
\label{eq:kernel-kaplan-meier}
\end{equation}
where, as before, $t_1<t_2<\cdots<t_m$ are the unique observed times in the training data. In practice, we add a tiny constant $\varepsilon>0$ to the denominator $n_K(t_{\ell}|x)$ to prevent division by~0; for simplicity, we omit writing this constant.
In equation~\eqref{eq:kernel-kaplan-meier}, the fraction
\begin{equation}
h_K(t_\ell|x):=\frac{d_K(t_\ell|x)}{n_K(t_\ell|x)}
\qquad\text{for }\ell=1,2,\dots,m
\label{eq:discrete-kernel-hazard}
\end{equation}
is a kernel estimate of the so-called (discrete-time) \textit{hazard function};
$h_K(t_\ell|x)$ is the estimated probability of a subject with feature vector~$x$ dying at time $t_\ell$ given that the subject has survived up to time $t_{\ell-1}$ (where $t_0:=0$). %
This kernel hazard estimate \eqref{eq:discrete-kernel-hazard} plays a crucial role in our proposed kernel learning method.

\subsection{Marginal and Local Prediction Intervals for Regression}
\label{sec:conformal-prediction}

To estimate \textit{marginal} and, separately, \textit{local} prediction intervals, we use split conformal prediction \citep{papadopoulos2002inductive,lei2015conformal} and its weighted variant \citep{tibshirani2019conformal}, respectively. For ease of exposition, we state these methods for the standard regression setting, where $(X_{1},Z_{1}),\dots,(X_{n},Z_{n})$ are i.i.d.~training data; we assume each feature vector $X_{i}\in\mathbb{R}^{d}$ is sampled from feature vector distribution $\mathbb{P}_{X}$ and each label $Z_{i}\in\mathbb{R}$ is sampled from a conditional distribution $\mathbb{P}_{Z|X=X_{i}}$. %
We aim to construct prediction intervals for predictions made using any regression algorithm~$\mathcal{A}$.

\paragraph{Split conformal prediction for regression}
Split conformal prediction assumes that to construct prediction intervals, we have access to a collection of $n_{\text{calib}}$ ``calibration'' data points $(X_{1}',Z_{1}'),\dots,(X_{n_{\text{calib}}}',Z_{n_{\text{calib}}}')$ independently sampled in the same way as the training data. Importantly, calibration data serve a different purpose than the usual validation data in machine learning: whereas validation data is used to help tune hyperparameters, calibration data cannot show up in the training procedure whatsoever.

Then to compute prediction intervals with coverage $1-\alpha$ for a user-specified tolerance $\alpha\in(0,1)$ and for any feature vector $x\in\mathbb{R}^{d}$, split conformal prediction does the following:
\begin{enumerate}[leftmargin=1.5em,itemsep=-0.25ex,topsep=1ex]
\item Use regression algorithm $\mathcal{A}$ with training data $(X_{1},Z_{1}),\dots,(X_{n},Z_{n})$ to estimate a regression function $\widehat{Z}$, i.e., $\widehat{Z}(x)$ is the predicted label value for feature vector $x$.
\item Compute residuals for the calibration data: $R_{i}=|Z_{i}'-\widehat{Z}(X_{i}')|$ for $i=1,\dots,n_{\text{calib}}$. We also include an additional residual value $R_{n_{\text{calib}}+1}:=\infty$.
\item Note that the residuals $R_{1},\dots,R_{n_{\text{calib}}+1}$ computed in step 2 form an empirical distribution on the real line augmented with $\{\infty\}$. Let $\widehat{q}$ be the $(1-\alpha)$-th quantile of this empirical distribution, i.e., if we denote the sorted residuals as $R_{(1)}\le R_{(2)}\le\cdots<R_{(n_{\text{calib}}+1)}=\infty$ (breaking ties randomly), then $\widehat{q}=R_{(\lceil(1-\alpha)(n_{\text{calib}}+1)\rceil)}$.
\item Output the prediction interval
$
\widehat{\mathcal{C}}^{\text{reg}}(x)=[\widehat{Z}(x)-\widehat{q},~\widehat{Z}(x)+\widehat{q}].
$
(The superscript stands for ``regression''.) We refer to $\widehat{q}$ as the ``radius'' of the interval.
\end{enumerate}
Adding a residual value of $\infty$ is so that if $\alpha$ is chosen to be extremely small (i.e., we demand the coverage $1-\alpha$ to be extremely close to 1), then the radius $\widehat{q}$ will be chosen to be $\infty$.

Importantly, the radius $\widehat{q}$ of $\widehat{\mathcal{C}}^{\text{reg}}(x)$ does \textit{not} depend on the test feature vector $x$, i.e., we estimate the same level of uncertainty for all $x$! This results from the fact that these prediction intervals are only valid marginally and not locally:
\begin{theorem}[Theorem 2.2 of~\citet{lei2018distribution}, first part]\label{thm:split-conformal}
Suppose that $(X_{n+1},Z_{n+1})$ is sampled independently the same way as the training data for the regression setup. Then
\[
\mathbb{P}\big(Z_{n+1}\in\widehat{\mathcal{C}}^{\text{reg}}(X_{n+1})\big)\ge1-\alpha.
\]
\end{theorem}
In the above guarantee, the probability is over randomness in sampling $X_{n+1}$ and \textit{not} conditioned on $X_{n+1}$ taking on a specific value. Put another way, the prediction intervals are valid averaged across test subjects, whose distribution is assumed to be the same as training subjects. Ideally, we want the level of uncertainty to depend on which test subject we look at. For example, we would like to construct a prediction interval $\widehat{\mathcal{C}}(x)$ such that
\[
\mathbb{P}\big(Z_{n+1}\in\widehat{\mathcal{C}}(x)\,\big|\,X_{n+1}=x\big)\ge1-\alpha.
\]
Unfortunately, obtaining guarantees for this setting is challenging; a series of impossibility results are provided by \citet{vovk2012conditional}, \citet{lei2014distribution}, and \citet{barber2019limits}.

\paragraph{Weighted split conformal prediction for regression}
Since conditioning exactly on $X_{n+1}=x$ is too much to ask for, recently \citet{tibshirani2019conformal} showed that with a relaxation, we can get valid prediction intervals using a specific notion of local coverage that relies on a kernel function $K$. For notational convenience, we now denote $x_{0}$ to be the test feature vector that we want this local coverage for. We construct a prediction interval $\widehat{\mathcal{C}}_K^{\text{reg}}(x;x_0)$ for any feature vector $x$ relative to how similar $x$ is to $x_{0}$ (according to kernel~$K$). The only change to the split conformal prediction procedure stated above is that in step 3, when we form the empirical distribution of the residuals, we instead form a \textit{weighted} empirical distribution; residual $R_{i}$ for calibration point $(X_{i}',Z_{i}')$ is assigned the weight $K(X_{i}',x_{0})$ for $i=1,2,\dots,n_{\text{calib}}$, and the inserted residual $R_{n_{\text{calib}}+1}=\infty$ is assigned the weight $K(x,x_{0})$. Put another way, residual $R_{i}$ is assigned the probability
\begin{equation}
p_{i}:=\begin{cases}
\frac{K(X_{i}',x_{0})}{\sum_{j=1}^{n_{\text{calib}}}K(X_{j}',x_{0})+K(x,x_{0})} & \text{if }i=1,2,\dots,n_{\text{calib}},\\
\frac{K(x,x_{0})}{\sum_{j=1}^{n_{\text{calib}}}K(X_{j}',x_{0})+K(x,x_{0})} & \text{if }i=n_{\text{calib}}+1.
\end{cases}
\label{eq:weighted-split-conformal-prediction-weights}
\end{equation}
We set the interval radius $\widehat{q}(x;x_0)$ to be the $1-\alpha$ quantile of this weighted empirical distribution, where as our notation suggests, the radius now depends on both $x$ and $x_0$.\footnote{Details on computing $\widehat{q}$: we first sort the residuals to obtain $R_{(1)}\le R_{(2)}\le\cdots\le R_{(n_{\text{calib}}+1)}=\infty$ (breaking ties randomly). Denote the assigned probabilities that correspond to these sorted residuals as $p_{(1)},p_{(2)},\dots,p_{(n_{\text{calib}}+1)}$. We then set $\widehat{j}$ to be the smallest index $j=1,2,\dots,n_{\text{calib}}+1$ such that $\sum_{i=1}^{j}p_{i}\ge1-\alpha$. Then we output $\widehat{q}(x;x_0)=R_{(\widehat{j})}$.} Step 4 is similar to before: $\widehat{\mathcal{C}}_{K}^{\text{reg}}(x;x_{0}) = [\widehat{Z}(x)-\widehat{q}(x;x_0),\widehat{Z}(x)+\widehat{q}(x;x_0)]$. We recover regular split conformal prediction when $K(x,x')=1$ for all feature vectors $x$ and $x'$, in which case the dependence on $x_0$ goes away, and $\widehat{q}(x;x_0)$ depends on neither $x$ nor $x_0$.

In what sense is this weighted version of split conformal prediction procedure ensuring local coverage? The idea is to slightly change how we sample feature vector $X_{n+1}$ compared to training data: instead of sampling $X_{n+1}$ from $\mathbb{P}_{X}$, we sample it from a version of $\mathbb{P}_{X}$ that has been weighted by kernel function $K(\cdot,x_{0})$. For simplicity, suppose that $\mathbb{P}_{X}$ has PDF $f_{X}$ (the theory works more generally even if $\mathbb{P}_{X}$ is, for example, a discrete distribution). Then we sample $X_{n+1}$ from a distribution with the following PDF parameterized by $x_0$:
\[
f_{X_{n+1}}(x;x_{0}):=\frac{K(x,x_{0})f_{X}(x)}{\int_{\mathbb{R}^{d}}K(x',x_{0})f_{X}(x')dx'}\qquad\text{for }x\in\mathbb{R}^{d}.
\]
For example, if we use the box kernel $K(x,x_{0})=\mathbf{1}\{\|x-x_{0}\|\le \sigma\}$, then $f_{X_{n+1}}(x)$ would be $f_{X}(x)$ restricted to have nonnegative probability whenever $x$ is within distance $\sigma$ of~$x_{0}$.  Aside from how $X_{n+1}$ is generated, we model label $Z_{n+1}$ to be generated using the same conditional distribution $\mathbb{P}_{Z|X}$ as for training data; i.e., $Z_{n+1}$ is sampled from $\mathbb{P}_{Z|X=X_{n+1}}$. We have the following guarantee:
\begin{theorem}[Equation (16) of \citet{tibshirani2019conformal}, rephrased]\label{thm:weighted-split-conformal}
We have
\[
\mathbb{P}\big(Z_{n+1}\in\widehat{\mathcal{C}}_{K}^{\text{reg}}(X_{n+1};x_{0})\,\big|\,X_{n+1}\sim f_{X_{n+1}}(\cdot;x_{0})\big)\ge1-\alpha.
\vspace{1em}
\]
\end{theorem}

\section{Deep Kernel Conditional Kaplan-Meier Estimator}
\label{sec:dksa}

We now present our method for learning a kernel function for the conditional Kaplan-Meier estimator \eqref{eq:kernel-kaplan-meier}. Recall from Section~\ref{sec:survival-analysis-problem-setup} that $t_1<t_2<\cdots<t_m$ are the unique observed times in the training data. Building on the work of \citet{brown1975use}, we minimize the following loss, which corresponds to maximizing the (mean) survival log-likelihood for the hazard function $h_K(t|x)$ in equation~\eqref{eq:discrete-kernel-hazard}:
\begin{align}
\text{loss }
L
:=-\frac{1}{n}
    \sum_{i=1}^n
      \bigg(
      &\delta_i \log[h_K(Y_i | X_i)]
      + (1 - \delta_i)
        \log[1 - h_K(Y_i | X_i)] \nonumber \\
      &+ \sum_{\ell=1}^m
          \ind\{t_\ell < Y_i\}
          \log[1 - h_K(t_\ell | X_i)]
      \bigg).
\label{eq:dksa-loss}
\end{align}
Note that \citet{brown1975use} did not use a kernel-based hazard function as we do; instead, Brown stated the above loss using a logistic hazard function $h(t_\ell|x):=\frac{1}{1+\exp(-\phi_\ell(x))}$ for an arbitrary parametric function $\phi:\mathbb{R}^d\rightarrow\mathbb{R}^m$, where $\phi(x)=(\phi_1(x),\phi_2(x),\dots,\phi_m(x))$. For this logistic hazard function, when $\phi$ is a neural net, we obtain the \mbox{\textsc{nnet-survival}} method of \citet{gensheimer2019scalable}.

By using the kernel-based hazard function $h_K(t|x)$ in equation~\eqref{eq:discrete-kernel-hazard}, we change Brown's loss to incorporate a kernel function $K$. Next, we parameterize the kernel function $K$ the same way as done by \citet{card2019deep} for kernel classification by setting
\begin{equation}
K(x,x'):=\exp(-\|\psi(x)-\psi(x')\|^2),
\label{eq:parametric-kernel}
\end{equation}
where $\psi$ is a user-specified base neural net. Put another way, we use a Gaussian kernel where the scaling factor that includes the variance is absorbed into the neural net $\psi$.

To summarize, the high-level idea is to minimize the loss $L$, which is a function of the kernel hazard function
\begin{align}
h_{K}(t_{\ell}|X_i)
&\;\overset{\text{equation }\eqref{eq:discrete-kernel-hazard}}{=}\;\frac{d_K(t_{\ell}|X_i)}{n_K(t_{\ell}|X_i)} \nonumber \\
&\;\overset{\text{equation }\eqref{eq:kernel-kaplan-meier-death-at-risk-quantities}}{=}\;
  \frac{\sum_{j=1}^n
          \delta_j K(X_i, X_j)\ind\{Y_j = t_{\ell}\}}
       {\sum_{j=1}^n
          K(X_i, X_j)\ind\{Y_j \ge t_{\ell}\}} \nonumber \\
&\;\overset{\text{equation }\eqref{eq:parametric-kernel}}{=}\;
  \frac{\sum_{j=1}^n
           \delta_j
           \exp(-\|\psi(X_i) - \psi(X_j)\|^2)
           \ind\{Y_j = t_{\ell}\}}
        {\sum_{j=1}^n
           \exp(-\|\psi(X_i) - \psi(X_j)\|^2)
           \ind\{Y_j \ge t_{\ell}\}}.
\label{eq:hurry-replace-me}
\end{align}
Thus, we minimize $L$ with respect to the parameters of the neural net $\psi$. After learning these parameters, we have thus learned the kernel function $K(x,x')=\exp(-\|\psi(x)-\psi(x')\|^2)$, which we plug into the conditional Kaplan-Meier estimator \eqref{eq:kernel-kaplan-meier} to produce an estimator $\widehat{S}(\cdot|x)$ of any subject's survival curve. We can then estimate subject-specific survival times using equation~\eqref{eq:median-surv-time}: $\widehat{T}(x):=\frac{1}{2}\big[\inf\{t\ge0 : \widehat{S}(t|x)\ge1/2\} + \sup\{t\ge0 : \widehat{S}(t|x)\le1/2\}\big]$.

Some implementation details are important for accurately estimating survival curves and also for scaling training to large datasets. Specifically, we (a) modify the loss with a leave-one-out strategy to avoid overfitting, (b) train with mini-batches to keep computation tractable, (c) further quantize time, and lastly (d) motivate some heuristics in how we choose an architecture for the neural net $\psi$. We describe these four pieces in detail next. The first two ideas are also used by \citet{card2019deep} for deep kernel classification, whereas the third idea is used by \citet{brown1975use} and more recently by \citet{lee2018deephit} in the \textsc{deephit} algorithm.

\paragraph{Leave-one-out strategy}
In the loss $L$, we form the kernel hazard function estimate $h_K(t_{\ell}|X_i)$ (at $\ell=1,2,\dots,m$) for the $i$-th training subject. To prevent overfitting, we disallow this estimate from using the $i$-th training subject's data. Thus, we replace $h_K(t_{\ell}|X_i)$ in equation~\eqref{eq:hurry-replace-me} with
\begin{align*}
h_{K \setminus i}(t_{\ell}|X_i)
&:=
  \frac{\sum_{j=1\text{ s.t.~}j\ne i}^n
           \delta_j
           \exp(-\|\psi(X_i) - \psi(X_j)\|^2)
           \ind\{Y_j = t_{\ell}\}}
        {\sum_{j=1\text{ s.t.~}j\ne i}^n
           \exp(-\|\psi(X_i) - \psi(X_j)\|^2)
           \ind\{Y_j \ge t_{\ell}\}}.
\end{align*}
\paragraph{Mini-batch learning}
To compute the $i$-th training subject's kernel hazard function estimate, we would have to compute the similarity of the $i$-th subject to the rest of the training subjects. Thus, computing the kernel hazard function estimates for all training subjects would require computation time that scales as $\mathcal{O}(n^2)$, which is prohibitively expensive. To scale training to large datasets, we use the standard approach of training in mini-batches so that the computation scales instead as $\mathcal{O}(b^2)$, where $b$ is the batch size.

\paragraph{Further quantizing time}
The loss $L$ sums over the unique observed times. For some datasets, the number of unique observed times $m$ can be large. We can further quantize the time grid and have the number of time points $m$ be a user-specified hyperparameter. In our experiments later (Section~\ref{sec:experiments-benchmark}), we either use all unique observed times (no quantization), or we set times $t_1<t_2<\cdots<t_m$ to be evenly spaced with $t_1$ and $t_m$ given by the minimum and maximum observed times in the training data. Note that quantization is not only for reducing computation time but can also affect accuracy of the estimated survival curves. In fact, for the datasets we consider, the running times are often roughly the same across quantization levels as we show in Appendix~\ref{sec:n-durations-cv-fitting-times}. Quantizing to fewer time points could be thought of as a form of regularization as we simplify the space of observed times.

\paragraph{Base neural net architecture choices and initialization}
There are many ways to choose the base neural net $\psi$. For example, one can even first train a different neural survival estimator and use its learned neural net (possibly with some final layers removed/modified) as an initial guess for $\psi$, which we then fine-tune by minimizing our kernel survival loss. However, to better understand how our approach works, we begin with simple shallow neural net architectures that are more interpretable before progressing to deeper networks. We then explain how any initial kernel function estimate, such as one learned using random survival forests \citep{ishwaran_2008}, can be used to warm-start the base neural net $\psi$.

Our simple heuristic neural net choices are inspired by existing work on kernel survival analysis by \citet{lowsky_2013} and \citet{chen2019nearest} that suggests that for some datasets, using Euclidean distance with standardized feature vectors and various standard kernel choices can already yield reasonable survival curve estimates. Thus, assuming that feature vectors are standardized, we can initialize $\psi$ close to or equal to identity. This means that $\psi$ is set to be a function mapping $\mathbb{R}^d$ to $\mathbb{R}^d$, where $d$ is the number of features.

The simplest choice we use for $\psi$ is $\psi_{\text{basic}}(x) := w x$, where the scalar $w\in\mathbb{R}$ is the only parameter. The resulting kernel function is $K(x,x')=\exp(-\|w x - w x'\|^2)={\exp(-w^2\|x-x'\|^2)}$, which is simply a Gaussian kernel with variance parameter $\sigma^2=2/w^2$. For training, we initialize $w$ to be 1. By choosing this neural net, we compare subjects using Euclidean distance in the original feature space with every feature equally weighted, and we are only learning a single variance parameter of a Gaussian kernel.

To allow for different features to have different weights, the next choice for~$\psi$ we use is %
\[
\psi_{\text{diag}}(x)
:= \begin{bmatrix}
        w_1 &   0 & \cdots &      0 \\
          0 & w_2 & \cdots &      0 \\
     \vdots &     & \ddots & \vdots \\
          0 &   0 & \cdots &    w_d
  \end{bmatrix} x,
\]
where $w=(w_1,\dots,w_d)\in\mathbb{R}^d$ is the parameter vector.
This choice for $\psi$ yields a \mbox{Gaussian} kernel with a diagonal covariance matrix, where the diagonal entries are $2/w_1^2,\dots,2/w_d^2$. The weights are initialized to all~1's. The learned weights indicate how much different features contribute to the Euclidean distance calculation; weights closer to 0 are considered less important in helping decide which subjects are similar.

To use deeper architectures while still initializing the base neural net to be close to identity, we import a key idea from highway \citep{srivastava2015training} and residual networks \citep{he2016deep} of letting the input be added to the output of another neural net. Let $\phi:\mathbb{R}^d\rightarrow\mathbb{R}^d$ be a user-specified, possibly deep neural net, and let $\xi$ be one of the simple choices we mentioned above ($\psi_{\text{basic}}$ or $\psi_{\text{diag}}$). %
Then we combine $\phi$ and $\xi$ via the following larger network
$\psi_{\text{residual}}(x;\phi,\xi):=\xi(x + \lambda \phi(x))$, where $\lambda>0$ is a hyperparameter. %

Lastly, we explain how, for any base neural net $\psi$, we can initialize it using any kernel function estimate, such as one learned using random survival forests \citep{ishwaran_2008}. Let $\widetilde{K}$ be the initial $n$-by-$n$ kernel matrix estimate for the $n$ training data, where entries of $\widetilde{K}$ are scaled to take on values between 0 and 1. For a trained random survival forest, $\widetilde{K}_{i,j}$ is given by the fraction of trees for which the $i$-th and $j$-th training data land in the same leaf. What we would like is for the neural net $\psi$ to satisfy the equation
\[
\widetilde{K}_{i,j} = \exp(-\|\psi(x_i) - \psi(x_j)\|^2),
\qquad\text{i.e.,}\qquad
\|\psi(x_i) - \psi(x_j)\|=\sqrt{\log(1/\widetilde{K}_{i,j})},
\]
where to prevent division by 0, we add a small constant to $\widetilde{K}_{i,j}$. To approximately achieve the above equality, we can use multidimensional scaling (MDS) \citep{borg2005modern} to learn an embedding $\widetilde{x}_1,\dots,\widetilde{x}_n\in\mathbb{R}^d$ such that $\|\widetilde{x}_i - \widetilde{x}_j\| \approx \sqrt{\log(1/\widetilde{K}_{i,j})}$ for all $i$ and $j$. Next, we warm-start the parameters of the neural net $\psi$ by minimizing the mean-squared error loss
\begin{equation}
\sum_{i=1}^n \sum_{j=1}^n \ind\{j>i\} \|\psi(x_i) - \widetilde{x}_i\|^2.
\label{eq:MSE-warm-start}
\end{equation}
In other words, we initialize $\psi$ by having it learn a mapping from the original feature space to the MDS embedding space, which is constructed to approximate Euclidean distances given by $\sqrt{\log(1/\widetilde{K}_{i,j})}$. Note that the MDS embedding dimension could be chosen to be smaller than the original feature dimension $d$ although we just use $d$ in our experiments later (matching the output dimension of our simple neural net choices from earlier).

\section{Prediction Intervals for Survival Time Estimates}
\label{sec:survival-conformal-prediction}

We now turn our attention to constructing marginally valid and locally valid prediction intervals for the survival time $T$ of a subject with feature vector $x$ using weighted split conformal prediction. Our exposition focuses on the weighted version since the standard unweighted version is a special case (when $K(x,x')=1$ for all feature vectors $x$ and $x'$). The rest of this section works with any kernel function $K$ and any estimator $\widehat{T}(x)$ of $T$ given $X=x$, where $\widehat{T}$ is learned using the training data $(X_{1},Y_{1},\delta_{1}),\dots,(X_{n},Y_{n},\delta_{n})$.

\paragraph{Constructing prediction intervals}
As with weighted split conformal prediction for regression, we assume that we have calibration data $(X_{1}',Y_{1}',\delta_{1}'),\dots,(X_{n_{\text{calib}}}',Y_{n_{\text{calib}}}',\delta_{n_{\text{calib}}}')$ sampled in the same way as the training data. To apply weighted split conformal prediction to survival time estimation, the two key ideas are that (a) weighted split conformal prediction works in the general setting when each data point's label is not just a real number but can also be the pair $(Y,\delta)$ consisting of a nonnegative observed time and a death indicator, and (b) earlier when we saw weighted split conformal prediction for regression, error was measured with the usual regression residuals, but more generally any function can be used to measure the ``error''; in conformal prediction literature, this ``error'' function is referred to as the \textit{nonconformity score}. While these two ideas are not new and already appear in various conformal prediction papers (e.g., \citealt{vovk2005algorithmic,shafer2008tutorial,vovk2012conditional}), to the best of our knowledge, they have not been applied to estimating subject-specific survival times, although they have been used to estimate prediction intervals for the conditional survival function $\widehat{S}(t|x)$ for a pre-specified time $t$ but only for random survival forests and that are only marginally valid \citep{bostr2017conformal}.

For survival time estimation, we use the following nonconformity score to measure the prediction error of $\widehat{T}$ for a data point $(x,y,\delta)$:
\begin{equation}
\mathcal{S}\big((x,y,\delta)\big)=\begin{cases}
|y-\widehat{T}(x)| & \text{if }\delta=1,\\
\max\{y-\widehat{T}(x),0\} & \text{if }\delta=0.
\end{cases}
\label{eq:survival-time-nonconformity}
\end{equation}
The intuition is that if $(x,y,\delta)$ is censored (i.e., $\delta=0$), then $y$ should be a lower bound on the survival time, so we incur no error if $\widehat{T}(x)\ge y$. Otherwise, if the point is not censored, then the error is the usual regression residual.

The changes to the weighted split conformal prediction method for regression from Section~\ref{sec:conformal-prediction} are as follows. First, instead of learning a regression function, we use training data to learn a survival time estimator $\widehat{T}$ in the inital step. Second, instead of regression residuals, we use the nonconformity score in equation~\eqref{eq:survival-time-nonconformity}. The last change is slightly more involved: the prediction ``interval'' gets replaced by a prediction set $\widehat{\mathcal{C}}_K^{\text{surv}}$, where we need to be able to check whether a label $(y,\delta)$ is inside $\widehat{\mathcal{C}}_K^{\text{surv}}$. For clarity of exposition, we explain this final change as part of the description of the algorithm.

We now state the weighted split conformal prediction procedure for survival time estimation, where we construct prediction sets $\widehat{\mathcal{C}}_K^{\text{surv}}(\cdot;x_0)$ local to test feature vector~$x_0$. In particular, for any subject with feature vector $x$, and any user-specified target coverage level $1-\alpha\in(0,1)$, note that $\widehat{\mathcal{C}}_K^{\text{surv}}(x;x_0)$ is the prediction set for $x$ accounting for how similar $x$ is to $x_0$. We construct the set $\widehat{\mathcal{C}}_K^{\text{surv}}(x;x_0)$ as follows:
\begin{enumerate}[leftmargin=1.5em,itemsep=-0.25ex,topsep=1ex]
\item Use training data $(X_{1},Y_{1},\delta_1),\dots,(X_{n},Y_{n},\delta_n)$ to learn a survival time estimator $\widehat{T}$.
\item Compute nonconformity scores for the calibration data using equation~\eqref{eq:survival-time-nonconformity}: $R_{i}=\mathcal{S}\big((X_i',Y_i',\delta_i')\big)$ for $i=1,\dots,n_{\text{calib}}$. We also include an additional score $R_{n_{\text{calib}}+1}:=\infty$.
\item Form a weighted empirical distribution for the scores $R_{1},\dots,R_{n_{\text{calib}}+1}$, where $R_i$ is assigned the probability given in equation~\eqref{eq:weighted-split-conformal-prediction-weights} and which we reproduce below:
\begin{equation*}
p_{i}:=\begin{cases}
\frac{K(X_{i}',x_{0})}{\sum_{j=1}^{n_{\text{calib}}}K(X_{j}',x_{0})+K(x,x_{0})} & \text{if }i=1,2,\dots,n_{\text{calib}},\\
\frac{K(x,x_{0})}{\sum_{j=1}^{n_{\text{calib}}}K(X_{j}',x_{0})+K(x,x_{0})} & \text{if }i=n_{\text{calib}}+1.
\end{cases}
\end{equation*}
Let $\widehat{q}(x;x_0)$ be the $(1-\alpha)$-th quantile of this weighted empirical distribution.
\item We output \textit{two} prediction intervals:
\begin{align*}
\widehat{\mathcal{C}}_K^{\text{observed}}(x;x_0)
&=[\widehat{T}(x)-\widehat{q}(x;x_0),~\widehat{T}(x)+\widehat{q}(x;x_0)], \nonumber \\
\widehat{\mathcal{C}}_K^{\text{censored}}(x;x_0)
&=[0,~\widehat{T}(x)+\widehat{q}(x;x_0)].
\end{align*}
Collectively, these two prediction intervals form the prediction set $\widehat{\mathcal{C}}_K^{\text{surv}}(x;x_0)$; specifically, to check whether any label $(y,\delta)$ is in $\widehat{\mathcal{C}}_K^{\text{surv}}(x;x_0)$, we first look at $\delta$. If $\delta=1$ (there's no censoring), then we check whether $y\in\widehat{\mathcal{C}}_K^{\text{observed}}(x;x_0)$; otherwise, we check whether $y\in\widehat{\mathcal{C}}_K^{\text{censored}}(x;x_0)$.\footnote{Technically, $\widehat{\mathcal{C}}_K^{\text{surv}}(x;x_0)=\big(\widehat{\mathcal{C}}_K^{\text{observed}}(x;x_0)\times\{1\}\big)\cup\big(\widehat{\mathcal{C}}_K^{\text{censored}}(x;x_0)\times\{0\}\big)$.} The intuition is that if $(y,\delta)$ is not censored, then the interval is just the usual regression interval. Otherwise, the prediction interval is for a censoring time, which can be any nonnegative value up to the survival time.
\end{enumerate}
We recover regular split conformal prediction for survival time estimation when ${K(x,x')=1}$ for all feature vectors $x$ and $x'$, in which case the dependence on $x_0$ disappears, $\widehat{q}$ depends on neither $x_0$ nor $x$, and we denote the resulting prediction set as $\widehat{\mathcal{C}}^{\text{surv}}(x)$.
The coverage guarantees are analogous to their regression counterparts (Theorems~\ref{thm:split-conformal} and~\ref{thm:weighted-split-conformal}):
\begin{proposition}\label{prop:coverage-guarantee}
(a) Suppose that $(X_{n+1},Y_{n+1},\delta_{n+1})$ is sampled independently the same way as the training data for survival analysis (given in Section~\ref{sec:survival-analysis-problem-setup}). Then
\[
\mathbb{P}\big((Y_{n+1},\delta_{n+1})\in\widehat{\mathcal{C}}^{\text{surv}}(X_{n+1})\big)\ge1-\alpha.
\]
(b) If instead $X_{n+1}$ is sampled from the distribution $f_{X_{n+1}}(x):=\frac{K(x,x_0)f_X(x)}{\int K(x',x_0)f_X(x')dx'}$ where $f_X$ is the PDF of feature vector distribution $\mathbb{P}_X$ (but $Y_{n+1}$ and $\delta_{n+1}$ are sampled in the same manner as training data conditioned on $X_{n+1}$), then
\[
\mathbb{P}\big((Y_{n+1},\delta_{n+1})\in\widehat{\mathcal{C}}_{K}^{\text{surv}}(X_{n+1};x_{0})\,\big|\,X_{n+1}\sim f_{X_{n+1}}(\cdot;x_{0})\big)\ge1-\alpha.\vspace{1em}
\]
\end{proposition}
Part (a) results from specializing the more general Proposition~4.1 of \citet{vovk2005algorithmic} to our survival analysis setup and our choice of nonconformity score. Part (b) uses the same proof as Theorem~2 of \citet{tibshirani2019conformal}, with the observation that the proof ideas still work if the label for each data point is of the form $(y,\delta)\in[0,\infty)\times\{0,1\}$.

\section{Numerical Experiments}
\label{sec:experiments}

We conduct experiments to understand (a) how well does our neural kernel survival analysis framework work in practice, (b) how well does the coverage guarantee of Proposition~\ref{prop:coverage-guarantee} hold in practice, and (c) how can the prediction intervals for survival times help us compare between different survival analysis methods. Our experiments use data on severely ill hospital patients from the Study to Understand Prognoses Preferences Outcomes and Risks of Treatment (\textsc{support}) \citep{knaus1995support} as well as three breast cancer datasets, which come from the Molecular Taxonomy of Breast Cancer International Consortium (\textsc{metabric}) \citep{curtis2012genomic}, the Rotterdam tumor bank (\textsc{rotterdam}) \citep{foekens2000urokinase}, and the German Breast Cancer Study Group (\textsc{gbsg}) \citep{schumacher_1994}. In all cases, the outcome of interest is time until death. We summarize some basic characteristics of these datasets in Table~\ref{tab:datasets}. Recent machine learning papers on survival analysis also test on these same datasets \citep{katzman2018deepsurv,kvamme2019time,kvamme2019continuous}. Our code is available at: \url{https://github.com/georgehc/dksa}

\begin{table}[!b]
\small
\centering
\begin{tabular}{|c|c|c|c|c|}
\hline
Dataset   & \# subjects  & \# features & \% censored & Observed times (min/median/max) \tabularnewline
\hline
\textsc{support}   & 8873 & 14 & 32.0\% & 0.10/7.59/66.70 months %
\tabularnewline
\textsc{metabric}  & 1904 &  9 & 42.1\% & 0/114.90/355.20 months \tabularnewline
\textsc{rotterdam} & 1546 &  7 & 37.4\% & 1.25/44.75/84 months \tabularnewline
\textsc{gbsg}      &  686 &  7 & 56.4\% & 0.26/35.61/87.36 months \tabularnewline
\hline
\end{tabular}
\caption{Basic characteristics of the survival datasets used.}
\label{tab:datasets}
\end{table}

\subsection{Benchmarking Deep Kernel Survival Analysis Against Existing Methods via Concordance Indices and Training Times}
\label{sec:experiments-benchmark}

For the \textsc{support} and \textsc{metabric} datasets, we use a random 70\%/30\% train/test split.  Following \citet{katzman2018deepsurv}, for the \textsc{rotterdam} and \textsc{gbsg} datasets, we train on \mbox{\textsc{rotterdam}} and test on \textsc{gbsg}. In each case, we use 5-fold cross-validation within training data to select different algorithms' hyperparameters (including neural net architecture choices); hyperparameter grids and details on neural net training are in Appendix~\ref{sec:hyperparameter-grids}. After selecting hyperparameters, we train on the full training data. We measure accuracy using the time-dependent concordance index (abbreviated as the $C^{\text{td}}$-index) by \citet{antolini2005time}. Roughly speaking, the $C^{\text{td}}$-index is the fraction of subjects correctly ordered by a survival curve prediction algorithm, accounting for time-dependent effects and censoring. It ranges in value from 0 to 1, with 1 being the highest score. We also record how long training each model takes during cross-validation.

We benchmark against two classical baselines---Cox proportional hazards \citep{cox_1972} and random survival forests \citep{ishwaran_2008}---as well as seven neural net baselines: \textsc{deepsurv} \citep{katzman2018deepsurv}, \textsc{deephit} \citep{lee2018deephit}, \textsc{mtlr} \citep{yu2011learning,fotso2018deep}, \textsc{nnet-survival} \citep{gensheimer2019scalable}, \mbox{\textsc{cox-cc}} \citep{kvamme2019time}, \textsc{cox-time} \citep{kvamme2019time}, and \textsc{pc-hazard} \citep{kvamme2019continuous}. The neural net approaches all depend on a base neural net $\phi$, which we take to be a multilayer perceptron (architecture details are in Appendix~\ref{sec:hyperparameter-grids}).

As for our neural kernel survival analysis approach (abbreviated \textsc{nks}), we experiment with several variants corresponding to different choices for the base neural net $\psi$ in equation~\eqref{eq:parametric-kernel}. Letting $\phi$ refer to a multilayer perceptron (same architecture choices as for the neural net baselines) and recalling our neural net architecture definitions in Section~\ref{sec:dksa}, we set the base neural net $\psi$ to $\psi_{\text{basic}}$, $\psi_{\text{diag}}$, $\psi_{\text{residual}}(\cdot; \phi, \psi_{\text{basic}})$, $\psi_{\text{residual}}(\cdot; \phi, \psi_{\text{diag}})$, and lastly~$\phi$; we refer to these five variants as \textsc{nks-basic}, \textsc{nks-diag}, \textsc{nks-res-basic}, \mbox{\textsc{nks-res-diag}}, and \textsc{nks-mlp}. Specifically for \textsc{nks-mlp}, we initialize neural net parameters via three strategies: standard neural net random initialization \citep{he2015delving}, random survival forests (the warm-start approach discussed at the end of Section~\ref{sec:dksa}), and \textsc{deephit} (warm-start using \textsc{deephit}'s neural net learned on the complete training data using the best hyperparameters found via cross-validation). Thus, accounting for the different initializations for \textsc{nks-mlp}, we test seven variants of \textsc{nks}. We include the final initialization with \textsc{deephit} as an illustrative example and, for simplicity, do not warm-start using the other neural baselines.

\begin{figure}[t]
\includegraphics[width=\linewidth]{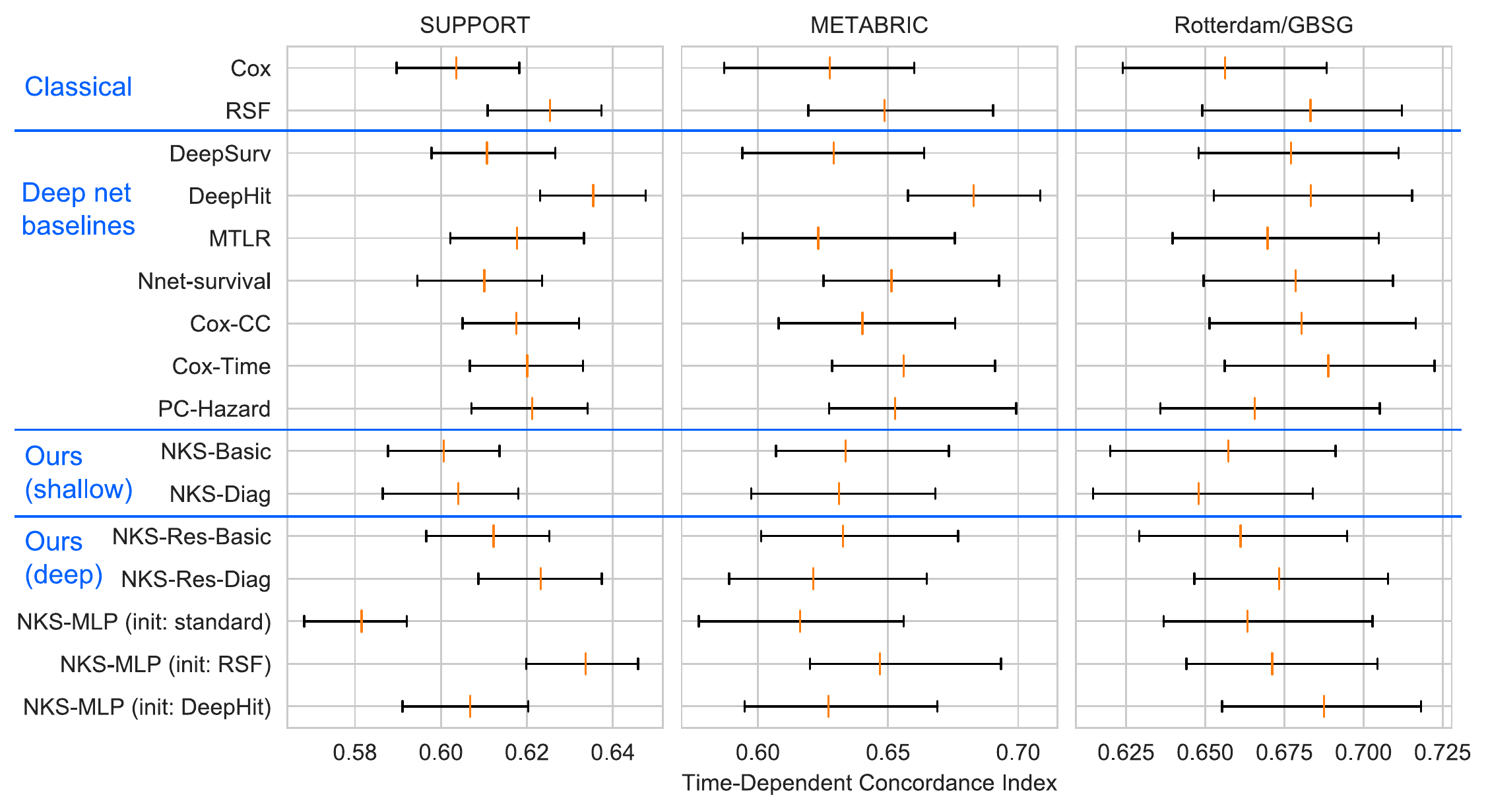}
\vspace{-2.5em}
\caption{Test set $C^{\text{td}}$-indices (vertical orange markers within intervals) across datasets and algorithms. Higher is better. Each interval is a 95\% bootstrap confidence interval.}
\label{fig:c-index}
\vspace{-1em}
\end{figure}

Test set $C^{\text{td}}$-indices are shown in Figure~\ref{fig:c-index} along with 95\% bootstrap confidence intervals (constructed by taking 100 bootstrap samples of the test data and then using the 2.5/97.5 percentiles). Among the baselines, we find that \textsc{deephit} consistently achieves the highest or nearly the highest $C^{\text{td}}$-indices, while random survival forests are competitive with many neural survival baselines. For our neural kernel survival estimators, the simplest variants \textsc{nks-basic} and \textsc{nks-diag} do not perform well although they are competitive with some baselines on the \textsc{metabric} dataset. Meanwhile, \textsc{nks-res-basic} and \textsc{nks-res-diag} tend to be more accurate than the simpler variants, with \textsc{nks-res-diag} competitive with multiple neural survival baselines across the datasets. As for the \textsc{nks-mlp} variants, we see that standard neural net initialization tends to result in noticeably worse accuracy than more cleverly initializing the neural net parameters with either random survival forests or \textsc{deephit}. With random survival forest initialization, \textsc{nks-mlp} tends to do better than all other \textsc{nks} variants tested, with the notable exception of \textsc{deephit} initialization leading to better performance on \textsc{rotterdam}/\textsc{gbsg}.

\begin{figure}[t]
\includegraphics[width=\linewidth]{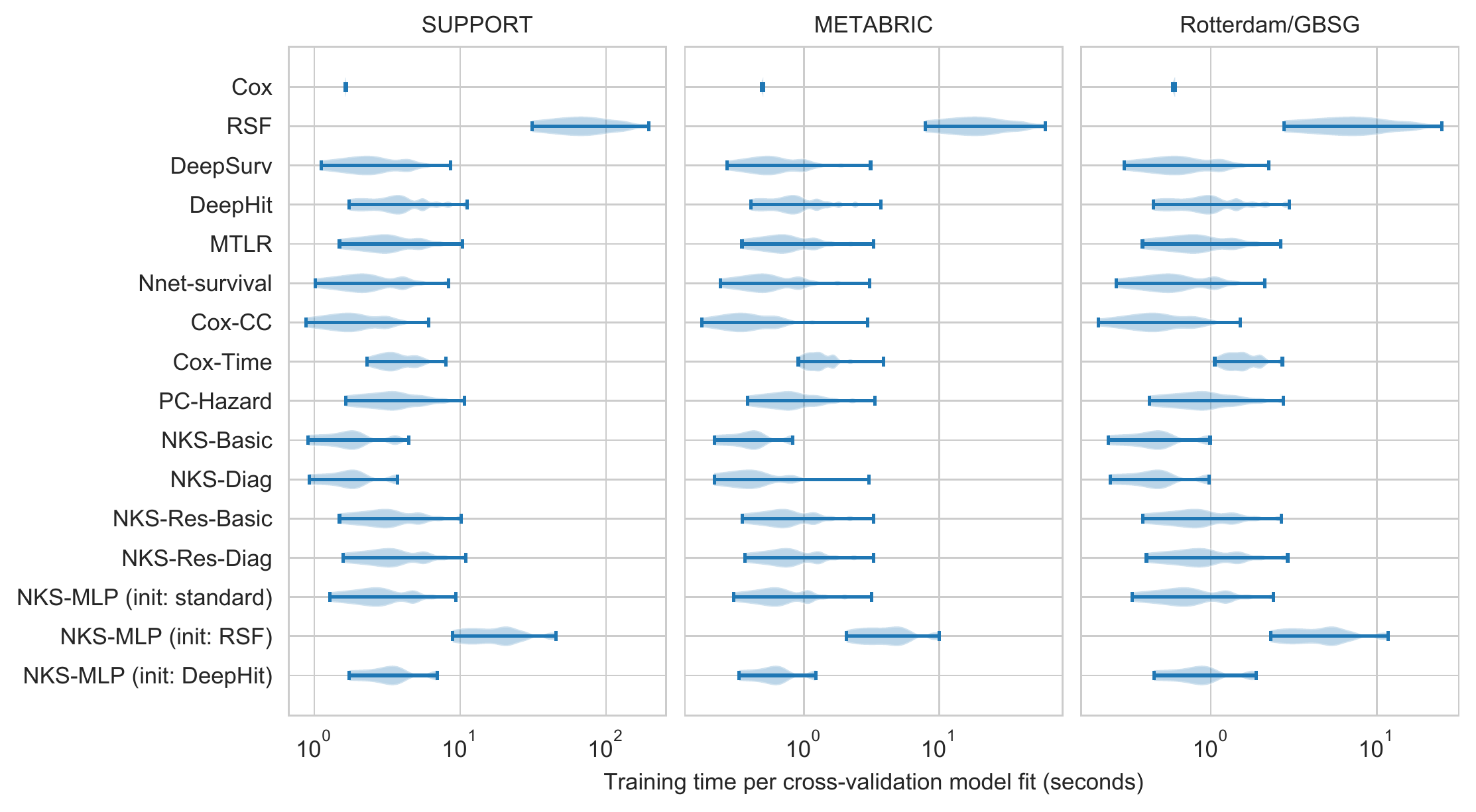}
\vspace{-2.5em}
\caption{Distributions of cross-validation model training times across datasets and algorithms. The horizontal axis is on a log scale. The times for \textsc{nks-mlp} with \textsc{rsf}/\textsc{deephit} initializations exclude training times of \textsc{rsf}/\textsc{deephit}.}
\label{fig:cv-fitting-times}
\vspace{-1em}
\end{figure}

Next, we give a sense of how long the different methods take to train. Since some methods have more hyperparameters than others and our hyperparameter search grids for different methods are chosen somewhat arbitrarily, rather than reporting how long the entirety of training (including cross-validation) takes, we instead report distributions of cross-validation model fitting times per method using violin plots as shown in Figure~\ref{fig:cv-fitting-times}. Note that the Cox proportional hazards model does not have hyperparameters, so there is no cross-validation step; however, to compare the Cox model's training time with other methods, we train it using data splits from 5-fold cross-validation strictly for the purposes of recording running times. Also, the reported times for \textsc{nks-mlp} variants initialized using random survival forests and \textsc{deephit} exclude the times needed to train the initializing algorithms. For the variant with random survival forest initialization, the times reported specifically are for warm-starting the neural net by minimizing loss~\eqref{eq:MSE-warm-start} and fine-tuning by minimizing the kernel hazard loss, where the fine-tuning takes time on par with \mbox{\textsc{nks-mlp}} using standard initialization.\footnote{We exclude the times for kernel matrix calculation and for finding an MDS embedding in the random survival forest warm-start since these two steps only need to be done once and do not depend on the neural net to be trained.} All algorithms are run on an Amazon Web Services \texttt{p3.2xlarge} instance (8 virtual CPU's on a Intel Xeon E5-2686v4 processor, 61 GiB RAM, and 1 NVIDIA Tesla V100 GPU with 16GiB memory). Overall, we find that the \textsc{nks} variants without \textsc{rsf}/\textsc{deephit} initialization have running times that are quite similar to the neural net baselines, with \textsc{nks-res-basic} and \textsc{nks-res-diag} having running times very similar to \textsc{deephit} and \textsc{mtlr}. %

\subsection{Examining Survival Time Prediction Intervals}
\label{sec:experiments-survival-conformal-prediction}

We now verify the statistical validity of our marginal and local prediction intervals, and we show how they can be used to compare survival analysis methods. For marginal prediction intervals, we use all survival estimators from the previous section, whereas for local prediction intervals, which require a kernel function, we only compare random survival forests with our \textsc{nks} variants. Our experiments here reuse the trained models from the previous section. In particular, we reuse the datasets' train/test splits but now treat test sets differently.

\begin{figure}[t]
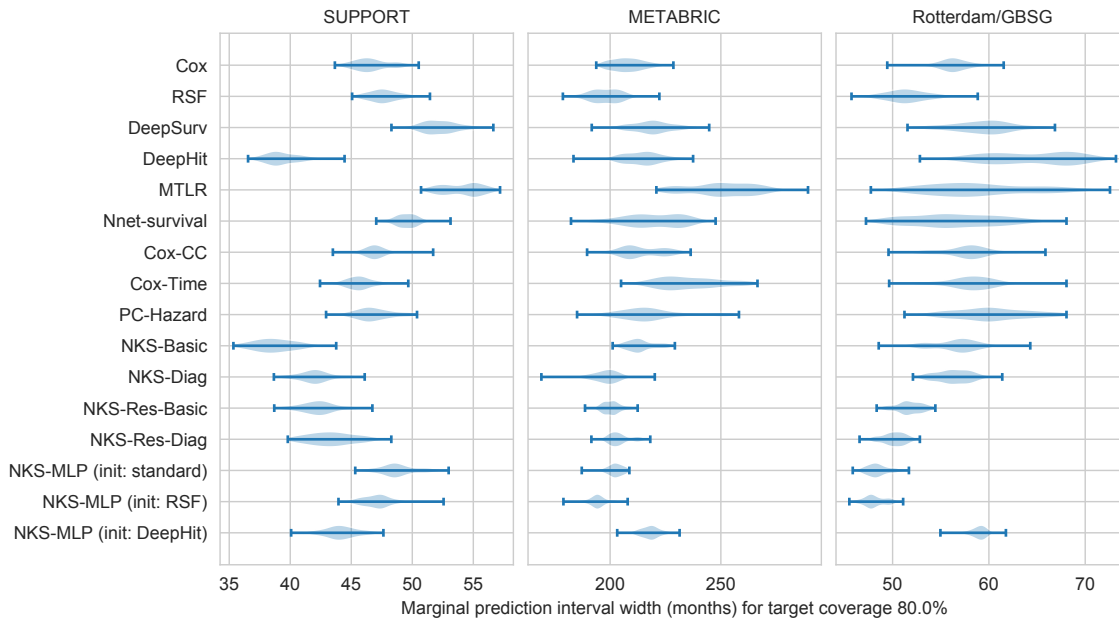

\includegraphics[width=\linewidth]{{{fig_interval_width_vs_methods_target0.8}}}
\vspace{-2.5em}
\caption{Distributions of marginal prediction interval widths at target coverage level ${1-\alpha}=0.8$. Smaller widths are better.}
\label{fig:marginal-interval-width}
\vspace{-1em}
\end{figure}

\paragraph{Marginal prediction intervals}
For each algorithm $\mathcal{A}$ we trained in Section~\ref{sec:experiments-benchmark} (using hyperparameters chosen via 5-fold cross-validation that only looks at the training data), and for different target levels $1-\alpha$, we conduct the following experiment:
\begin{enumerate}[leftmargin=1.5em,itemsep=-0.25ex,topsep=1ex]
\item Randomly divide the test set into two halves, one to treat as calibration data for constructing prediction intervals and one to treat as the \textit{proper} test data.
\item (Split conformal prediction) Algorithm $\mathcal{A}$ yields a conditional survival function estimate $\widehat{S}(t|x)$, from which we obtain a survival time estimator $\widehat{T}(x)$. Using the calibration data, compute the radius $\widehat{q}$ of prediction intervals; recall that this radius does not depend on which test point we evaluate at later.
\item For every proper test data point $(x,y,\delta)$, we check whether $(y,\delta)\in\widehat{\mathcal{C}}^{\text{surv}}(x)$.
\item Record the fraction of proper test points that fall in the constructed prediction intervals; this fraction is the \textit{empirical coverage probability}. Also record the prediction interval width $2 \widehat{q}$.
\end{enumerate}
We repeat the above experiment 100 times for different calibration/proper test splits. Thus, for each dataset/algorithm/target coverage level, we have a distribution of 100 empirical coverage probabilities, and a distribution of 100 prediction interval widths. For target coverage level $1-\alpha=0.8$, we report the means and standard deviations of empirical coverage probabilities in Table~\ref{tab:marginal-PI-coverage} and display distributions of prediction interval widths as violin plots in Figure~\ref{fig:marginal-interval-width}.

\begin{table}[!t]
\footnotesize
\centering
\begin{tabular}{|c|c|c|c|}
\cline{2-4} 
\multicolumn{1}{c|}{} & \textsc{support} & \textsc{metabric} & \textsc{rotterdam}/\textsc{gbsg}\tabularnewline
\hline
\textsc{cox} & $0.802\pm0.015$ & $0.807\pm0.032$ & $0.804\pm0.029$ \tabularnewline
\hline
\textsc{rsf} & $0.802\pm0.015$ & $0.807\pm0.035$ & $0.806\pm0.031$ \tabularnewline
\hline
\textsc{deepsurv} & $0.802\pm0.014$ & $0.807\pm0.033$ & $0.803\pm0.029$ \tabularnewline
\hline
\textsc{deephit} & $0.802\pm0.017$ & $0.805\pm0.038$ & $0.803\pm0.032$ \tabularnewline
\hline
\textsc{mtlr} & $0.802\pm0.015$ & $0.803\pm0.036$ & $0.803\pm0.028$ \tabularnewline
\hline
\textsc{nnet-survival} & $0.802\pm0.014$ & $0.811\pm0.033$ & $0.804\pm0.030$ \tabularnewline
\hline
\textsc{cox-cc} & $0.803\pm0.014$ & $0.806\pm0.034$ & $0.804\pm0.028$ \tabularnewline
\hline
\textsc{cox-time} & $0.803\pm0.015$ & $0.811\pm0.031$ & $0.802\pm0.030$ \tabularnewline
\hline
\textsc{pc-hazard} & $0.801\pm0.014$ & $0.807\pm0.035$ & $0.804\pm0.031$ \tabularnewline
\hline
\textsc{nks-basic} & $0.801\pm0.017$ & $0.807\pm0.033$ & $0.806\pm0.030$ \tabularnewline
\hline
\textsc{nks-diag} & $0.802\pm0.017$ & $0.805\pm0.036$ & $0.806\pm0.028$ \tabularnewline
\hline
\textsc{nks-res-basic} & $0.803\pm0.017$ & $0.805\pm0.032$ & $0.808\pm0.029$ \tabularnewline
\hline
\textsc{nks-res-diag} & $0.802\pm0.018$ & $0.806\pm0.031$ & $0.806\pm0.029$ \tabularnewline
\hline
\textsc{nks-mlp} & $0.802\pm0.017$ & $0.803\pm0.033$ & $0.805\pm0.031$ \tabularnewline
\hline
\textsc{nks-mlp} (init: \textsc{rsf}) & $0.802\pm0.015$ & $0.807\pm0.032$ & $0.805\pm0.031$ \tabularnewline
\hline
\textsc{nks-mlp} (init: \textsc{deephit}) & $0.802\pm0.017$ & $0.803\pm0.037$ & $0.806\pm0.030$ \tabularnewline
\hline 
\end{tabular}
\caption{Empirical coverage probabilities of marginal prediction intervals (mean $\pm$ std dev) at target coverage level $1-\alpha=0.8$. As desired, all values are close to 0.8.}
\label{tab:marginal-PI-coverage}
\end{table}

\begin{figure}[t]
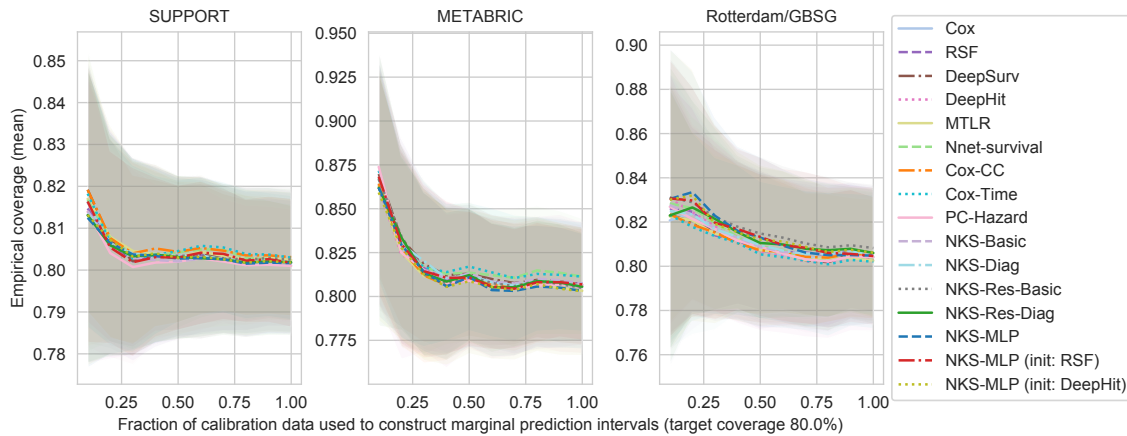

\includegraphics[width=\linewidth]{{{fig_marginal_emp_coverage_vs_calib_frac_target0.8}}}
\vspace{-2.5em}
\caption{Empirical coverage probability (mean $\pm$ std dev) vs fraction of calibration data used to construct the marginal prediction intervals at target coverage level $1-\alpha=0.8$. Because the error bars (shaded with colors corresponding to algorithms) heavily overlap, they appear gray.}
\label{fig:marginal-emp-coverage-vs-calib-frac}
\vspace{-1em}
\end{figure}

As shown in Table~\ref{tab:marginal-PI-coverage}, when constructing prediction intervals with a user-specified target coverage level of~0.8, all the empirical coverage probabilities are indeed close to~0.8. Varying the target coverage from 0.5 to 0.95, we found that the same patterns holds in all cases, so we omit the tables for these other coverage levels. Instead, we examine how the empirical coverage probabilities differ when we use less calibration data by varying the amount of calibration data from 10\% to 100\% of the full calibration set described above, leaving the proper test dataset size fixed. For target coverage level 0.8, we plot the empirical coverage probability vs the amount of calibration data used in Figure~\ref{fig:marginal-emp-coverage-vs-calib-frac}. We see that with very little calibration data, the empirical coverage probabilities tend to be higher than the true target coverage, but as the amount of calibration data increases, the empirical coverage curves slope downward and then flatten out, converging to the true target coverage level. Once again, we get similar plots for other target coverage levels, so we omit these other plots.

\begin{figure}[p]
\centering
\includegraphics[width=\linewidth]{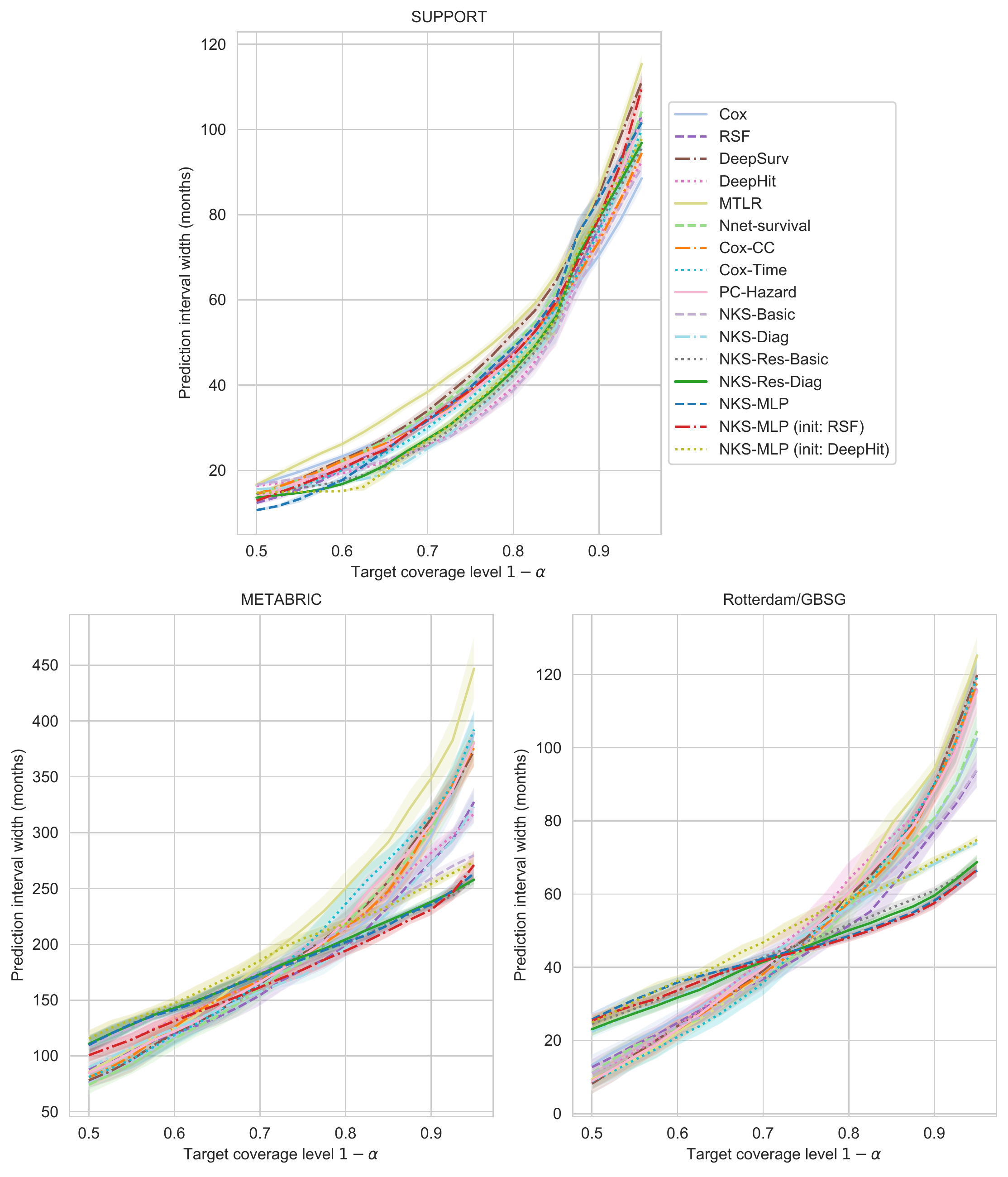}
\vspace{-2em}
\caption{Marginal prediction interval width (mean $\pm$ 1 std dev) vs target coverage level $1-\alpha$ across datasets and algorithms. The legend is for all plots. To give a relative sense of whether the interval widths are too wide, recall from Table~\ref{tab:datasets} that the maximum survival times (months) are 66.70 for \textsc{support}, 355.20 for \textsc{metabric}, and 87.36 for \textsc{gbsg} (\textsc{rotterdam} is used only as training data).}
\label{fig:support-qhat-curves}
\end{figure}

Now that we have established that with enough calibration data, the empirical coverage probabilities for marginal prediction intervals are close to target coverage levels, we return to using the full calibration set and examine the prediction intervals' widths $2\widehat{q}$. Importantly, which survival analysis method has the smallest interval width varies by dataset and also by the target coverage level. We plot the mean interval width vs the target coverage level $1-\alpha$ across datasets and methods in Figure~\ref{fig:support-qhat-curves}. We see that for the \textsc{support} dataset, for target coverage levels 0.75--0.85, \textsc{nks-basic} and \textsc{deephit} have the smallest interval widths, whereas at higher target coverage levels 0.9--0.95, the Cox model has the smallest interval widths. For the \textsc{metabric} dataset, \textsc{nks-mlp} with \textsc{rsf} initialization has the smallest interval widths at target coverage levels 0.8--0.9, and at higher target coverage levels 0.9--0.95, \mbox{\textsc{nks-res-basic}}, \textsc{nks-res-diag}, and \textsc{nks-mlp} (standard and \textsc{rsf} initializations) have the smallest interval widths. For \textsc{rotterdam}/\textsc{gbsg}, we find that for target coverage levels 0.8--0.95, \textsc{nks-mlp} (standard and \textsc{rsf} initializations) have the smallest interval widths. Overall, \textsc{nks} variants are able to achieve among the smallest interval widths for a variety of target coverage levels.

\paragraph{Local prediction intervals}
To verify the validity of local prediction intervals, we modify the experiment we conduct for marginal prediction intervals. Note that now we only use methods that learn a kernel and specifically experiment with our \textsc{nks} variants along with random survival forests. For different datasets and different target coverage levels $1-\alpha$, we run the following experiment:
\begin{enumerate}[leftmargin=1.5em,itemsep=-0.25ex,topsep=1ex]
\item Randomly divide the test set into two halves, one to treat as calibration data for constructing prediction intervals and one to treat as the proper test data.
\item Randomly sample (with replacement) 100 proper test points that we shall construct local confidence intervals with respect to; denote this list of 100 points as $\mathcal{X}_{\text{local-centers}}$.
\item For each point $x_0\in\mathcal{X}_{\text{local-centers}}$:
\begin{enumerate}[leftmargin=1.5em,itemsep=-0.25ex,topsep=0ex]
\item Randomly sample (with replacement) 100 proper test points, where the probability of sampling each point $x$ is weighted proportional to $K(x,x_0)$; denote this list of 100 points as $\mathcal{X}_{\text{subjects-similar-to-}x_0}$.
\item For each point $x\in\mathcal{X}_{\text{subjects-similar-to-}x_0}$, we check whether the point's true label $(y,\delta)$ is in $\widehat{\mathcal{C}}_K^{\text{surv}}(x;x_0)$. Also record the interval radius $\widehat{q}(x;x_0)$.
\item Record the fraction of points in $\mathcal{X}_{\text{subjects-similar-to-}x_0}$ that land in their respective local prediction intervals in the previous step. This fraction is the empirical coverage probability.
\end{enumerate}
\end{enumerate}
We repeat the above experiment 100 times for different random calibriation/proper test splits. For target coverage level $1-\alpha=0.8$, we report means and standard deviations of empirical coverage probabilities in Table~\ref{tab:local-PI-coverage}. At other target coverage levels, the empirical coverage probabilities again are close to the desired target coverage levels; we omit these additional tables.

\begin{figure}[p]
\centering
\includegraphics[width=\linewidth]{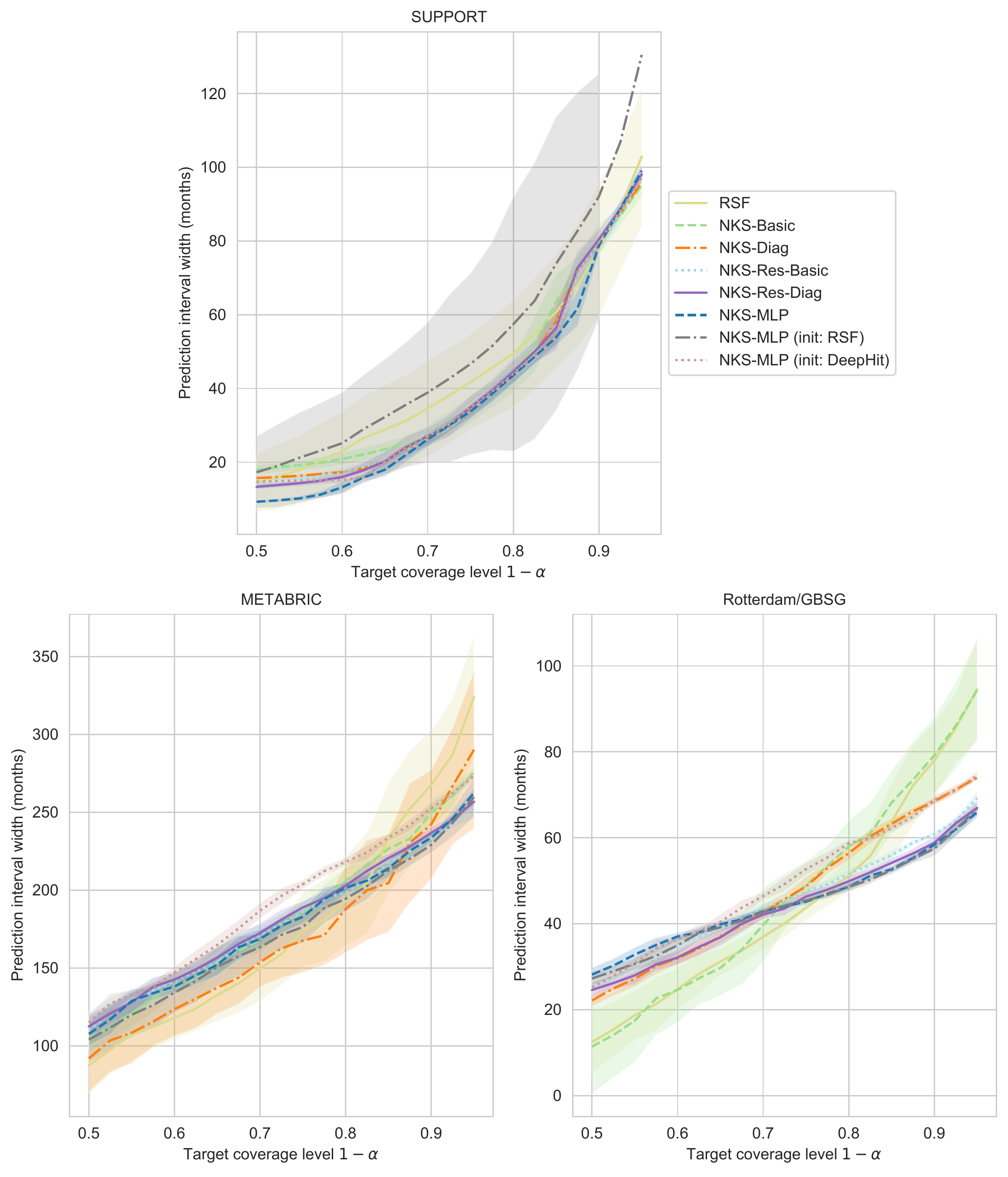}
\vspace{-2em}
\caption{Local prediction interval width (median $\pm$ quartile deviation) vs target coverage level $1-\alpha$ across datasets and algorithms. The legend is for all plots. To give a relative sense of whether the interval widths are too wide, recall from Table~\ref{tab:datasets} that the maximum survival times (months) are 66.70 for \textsc{support}, 355.20 for \textsc{metabric}, and 87.36 for \textsc{gbsg} (\textsc{rotterdam} is used only as training data). The top plot's gray error bars stop when the quartile deviation becomes infinite.}
\label{fig:local-qhat-curves}
\end{figure}

This time around, we do not report means and standard deviations of the recorded prediction interval widths since sometimes these can be infinity, so the average is not defined. The reason is simple: for different subjects, we have different uncertainties about their predicted survival times \emph{relative to how similar they are to specific other subjects}, and sometimes we do have prediction intervals of infinite width to indicate extremely high uncertainty at the desired target coverage level $1-\alpha$. Instead of means and standard deviations of prediction interval widths, we could use medians and quartile deviations (half of the interquartile range). We plot local prediction interval width vs target coverage level $1-\alpha$ across datasets and methods in Figure~\ref{fig:local-qhat-curves}. For the \textsc{support} dataset, nearly all \textsc{nks} variants except for \textsc{nks-mlp} with \textsc{rsf} initialization have as small or smaller interval widths than \textsc{rsf}. For \textsc{metabric} and \textsc{rotterdam}/\textsc{gsbg} datasets, at lower target coverage levels, \textsc{rsf} can achieve among the smallest interval widths but at higher target coverage levels, the deep \textsc{nks} variants start achieving the smallest interval widths.

\begin{table}[!t]
\footnotesize
\centering
\begin{tabular}{|c|c|c|c|}
\cline{2-4} 
\multicolumn{1}{c|}{} & \textsc{support} & \textsc{metabric} & \textsc{rotterdam}/\textsc{gbsg}\tabularnewline
\hline 
\textsc{rsf} & $0.810\pm0.083$ & $0.802\pm0.077$ & $0.803\pm0.060$ \tabularnewline
\hline
\textsc{nks-basic} & $0.817\pm0.064$ & $0.806\pm0.053$ & $0.816\pm0.079$ \tabularnewline
\hline
\textsc{nks-diag} & $0.802\pm0.043$ & $0.823\pm0.085$ & $0.805\pm0.050$ \tabularnewline
\hline
\textsc{nks-res-basic} & $0.802\pm0.045$ & $0.804\pm0.053$ & $0.805\pm0.052$ \tabularnewline
\hline
\textsc{nks-res-diag} & $0.802\pm0.045$ & $0.803\pm0.053$ & $0.806\pm0.052$ \tabularnewline
\hline
\textsc{nks-mlp} & $0.801\pm0.044$ & $0.804\pm0.055$ & $0.804\pm0.051$ \tabularnewline
\hline
\textsc{nks-mlp} (init: \textsc{rsf}) & $0.849\pm0.090$ & $0.804\pm0.058$ & $0.801\pm0.051$ \tabularnewline
\hline
\textsc{nks-mlp} (init: \textsc{deephit}) & $0.801\pm0.043$ & $0.804\pm0.050$ & $0.804\pm0.050$ \tabularnewline
\hline 
\end{tabular}
\caption{Empirical coverage probabilities of local prediction intervals (mean $\pm$ std dev) at target coverage level $1-\alpha=0.8$. As desired, all values are close to 0.8.}
\label{tab:local-PI-coverage}
\end{table}

\section{Discussion and Limitations}

\paragraph{Deep kernel survival analysis}
We have presented a new neural net framework for learning kernel functions for kernel survival analysis. This framework minimizes a survival loss to learn a kernel function and can easily be extended: for example, we can add regularization, explore base neural nets that account for other structure (e.g., recurrent neural nets for temporal data), and experiment with a wide array of optimizers. In contrast, the only existing approach for automatically learning a kernel for survival analysis without choosing from a collection of pre-specified kernels is to use random survival forests, which do not have a known global objective function that is minimized. As we have demonstrated, random survival forests can actually be used to warm-start neural kernel learning.

For simplicity, the survival loss we use is based on the likelihood specified by \citet{brown1975use}. Other survival loss functions are also possible. For example, the kernel hazard function $h_K(t_{\ell}|x)$ (given in equation~\eqref{eq:discrete-kernel-hazard}) can readily be converted to a survival time probability mass function instead (see the derivation in Section 3.1 of~\citet{kvamme2019continuous}), which can then be directly used in the \textsc{deephit} loss function. Conceptually, this amounts to using the loss function that we are already using with an additional ranking loss term tailored toward optimizing the concordance index.

Stepping away from estimating survival curves altogether, we remark that the idea by \citet{card2019deep} of parameterizing the kernel function as a neural net that we used in our neural survival analysis approach can also be combined with survival support vector machines \citep{shivaswamy2007support,khan2008support} to directly estimate survival times. Thus, it is possible to automatically learn a kernel for predicting different subjects' survival times without ever estimating their survival curves.

\paragraph{Subject-specific prediction intervals}
We have also shown how to construct prediction intervals for subject-specific survival times, where we produce intervals that are marginally valid (averaged across test population) and, separately, intervals that are locally valid (averaged across subjects similar to a specific individual). These intervals depend on a user-specified target coverage level, which are like confidence levels for confidence intervals: if we demand a higher target coverage level (e.g., 0.99), then the resulting intervals are wider. %

Both types of intervals enable benchmarking survival time estimators by their prediction interval widths at different target coverage levels: marginal prediction intervals can be used for all survival time estimators, whereas local prediction intervals require a kernel function to be specified. We remark that for local prediction intervals, the kernel function is only needed after the survival time estimator has been trained. For example, we can produce locally valid prediction intervals for a survival estimator that does not use a kernel function if, after training it, we separately either manually specify or automatically learn a kernel function strictly for the purposes of interval construction.

Prediction intervals give us a way to more carefully choose which survival estimator we should be using. For example, suppose that at a target coverage level of 0.9, all prediction algorithms under consideration yield prediction interval widths that are far too wide to be practically useful. Then we know that we have to settle for a lower target coverage level, since lower target coverage levels correspond to narrower prediction intervals. As we have seen in the numerical experiments, at different target coverage levels, which survival estimators have the narrowest prediction intervals varies. Put another way, much like how different estimators have different bias-variance tradeoffs, they also have different prediction interval width vs target coverage level tradeoffs.

We suspect locally valid prediction intervals to be more useful in practice if we care about individual-specific prediction and clinical decision support. For example, using a kernel survival analysis method, we can predict the survival time of a specific test subject. Using the kernel function, we can then identify the training subjects most similar to the test subject. We can then examine what the local prediction intervals are for the test subject relative to each of these most similar training subjects. The different local prediction intervals can vary in width and enable us to gauge prediction uncertainty specific to the test subject. %

Our work has a number of limitations. We highlight a few of them below.

\paragraph{Computation}
The datasets we tested on are relatively small, so the computation times for both training and testing using \textsc{nks} variants were on par with various deep net baselines. However, our approach inherently does not scale well at test time to substantially larger datasets due to the need to compute distances between test data and all training data. We can accelerate this computation using, for instance, approximate nearest neighbor search in Euclidean space (since we map each point to an embedding space via the base neural net~$\psi$ and compare embedded points via Euclidean distance), or using random Fourier features for approximating Gaussian kernels \citep{rahimi2008random}. The latter could also be used to enable mini-batch neural kernel training with very large batch sizes.

\paragraph{Accuracy}
In terms of $C^{\text{td}}$-indices, deep kernel survival estimators \textsc{nks-res-diag} and \textsc{nks-mlp} with random survival forest initialization are competitive with many baselines. However, none of the survival analysis methods tested achieve a \mbox{$C^{\text{td}}$-index} close to 1 on any of the datasets. Moreover, for all datasets, deep net approaches can be competitive with but for the most part do not significantly outperform random survival forests. Even in comparison to the Cox model, the increase in $C^{\text{td}}$-index by using a deep learning approach might not be justified in a clinical application when accounting for the loss in model interpretability. Perhaps on much larger survival datasets, we could see more dramatic gains from deep learning vs the Cox and random survival forest baselines.

For neural kernel estimators, we suspect that different base neural net choices and initializations are needed to guide learning compared to neural net approaches that are not kernel-function-based. %
For example, initializing \textsc{nks-mlp} using either standard neural net initialization or \textsc{deephit} did not tend to work as well as using random survival forest initialization, which might be due to random survival forests being related to kernel learning. The only other base neural nets we experimented with are slight perturbations of the identity function. Further investigation is needed to understand the landscape of neural net architectures and random initialization strategies that are highly effective for learning kernel functions.

\paragraph{Reducing uncertainty}
Lastly, we remark that our prediction intervals, while statistically valid, still have widths that are quite wide. It is unclear to us what realistic assumptions we could incorporate to shrink these intervals while maintaining statistical validity. %
Separately, a future research direction could look at whether we can learn survival estimators that focus on getting marginal prediction intervals as narrow as possible for a user-specified band of intermediate target coverage levels, allowing for such an estimator to have arbitrarily wide intervals above the user-specified band.

\acks{The author thanks the anonymous reviewers for very helpful feedback.}

\bibliography{dksa}

\appendix

\section{Estimating Subject-Specific Survival Times}
\label{sec:surv-time-estimation}

Survival time estimation is a well-studied problem in survival analysis with standard solutions that are based on having already computed a conditional survival function estimate $\widehat{S}(t|x)$. The median survival time estimator~\eqref{eq:median-surv-time} that we use is a slight modification of the original one suggested by \citet{reid1981estimating}: $\widehat{T}(x)=\inf\{t\ge0 : \widehat{S}(t|x)\ge1/2\}$. The intuition for these median survival time estimators comes from observing that $S(t|x)$ is 1 minus the CDF of the distribution $\mathbb{P}_{T|X}$, and that where a CDF crosses 1/2 corresponds to a median of the distribution. Our modification of Reid's original estimator just uses the idea that in computing medians, a standard approach is to average the two closest values to the 50th percentile rather than only using one of the values, although it is possible for these two closest values to coincide. As a toy example of this idea, when computing the median of a sequence of numbers, if the sequence is of even length, we sort the values and average the two values that are in the middle.

An alternative to using a median survival time estimate is to instead have $\widehat{T}(x)$ estimate $\mathbb{E}[T|X=x]$. To do this, first recall that for any nonnegative random variable $Z$, we have $\mathbb{E}[Z]=\int_{0}^{\infty}\mathbb{P}(Z>t)dt$. Then with the choice $Z=(T|X=x)$,
\[
\mathbb{E}[T|X=x]=\int_{0}^{\infty}\mathbb{P}(T>t|X=x)dt=\int_{0}^{\infty}S(t|x)dt.
\]
Thus, we can estimate $T$ given $X=x$ with the estimator $\widehat{T}(x):=\int_{0}^{\infty}\widehat{S}(t|x)dt$, where we use numerical integration such as the trapezoidal rule. %

\section{Hyperparameter Grids and Neural Net Training Details}
\label{sec:hyperparameter-grids}

For random survival forests, we fix the number of trees to be 100 and search over the following hyperparameters:
\begin{itemize}[leftmargin=1.5em,itemsep=-0.25ex,topsep=1ex]
\item Maximum features per split: 2, 4, 6
\item Minimum training samples per leaf: 8, 32, 128
\end{itemize}
For all neural net methods, we train with the Adam optimizer \citep{kingma2014adam} searching over the following hyperparameters:
\begin{itemize}[leftmargin=1.5em,itemsep=-0.25ex,topsep=1ex]
\item Number of epochs: 10, 20
\item Batch size: 64, 128
\item Learning rate: 0.01, 0.001
\end{itemize}
For methods that work on a discretized time grid including our \textsc{nks} variants, we search over the number of time points $m=64$ and $m=128$.

The neural survival analysis baselines as well as \textsc{nks-mlp}, \textsc{nks-res-basic}, and \textsc{nks-res-diag} all depend on a base neural net $\phi$, which we take to be a multilayer perceptron. We search over the following grid for this multilayer perceptron:
\begin{itemize}[leftmargin=1.5em,itemsep=-0.25ex,topsep=1ex]
\item Number of hidden layers: 1, 2, 4
\item Number of nodes per hidden layer: 16, 32, 64
\end{itemize}
We set the hidden layers to all use ReLU activation followed by BatchNorm \citep{ioffe2015batch}. The final fully-connected output layer has a number of output nodes that depends on the neural survival analysis used. \textsc{deepsurv}, \textsc{cox-cc}, and \textsc{cox-time} all require the output of $\phi$ to be a single number that has no bias added (the bias would get folded into the baseline hazard anyways), while \textsc{deephit}, \textsc{nnet-survival}, and \textsc{pc-hazard} allow a bias but require the number of output nodes to be equal to the number of discrete time steps~$m$. As we mentioned in Section~\ref{sec:dksa}, for simplicity, we constrain our \textsc{nks} variants to have the number of output nodes be the same as the number of input features $d$.

For \textsc{nks-res-basic} and \textsc{nks-res-diag}, we set the hyperparameter $\lambda$ to be 0.1 (recall that the neural net we use for these two methods are $x\mapsto\psi_{\text{basic}}(x+\lambda\phi(x))$ and $x\mapsto\psi_{\text{diag}}(x+\lambda\phi(x))$) as to intentionally bias the initial network to be close to identity. %

\section{Training Times for Different Time Grid Discretizations}
\label{sec:n-durations-cv-fitting-times}

\begin{figure}
\includegraphics[width=\linewidth]{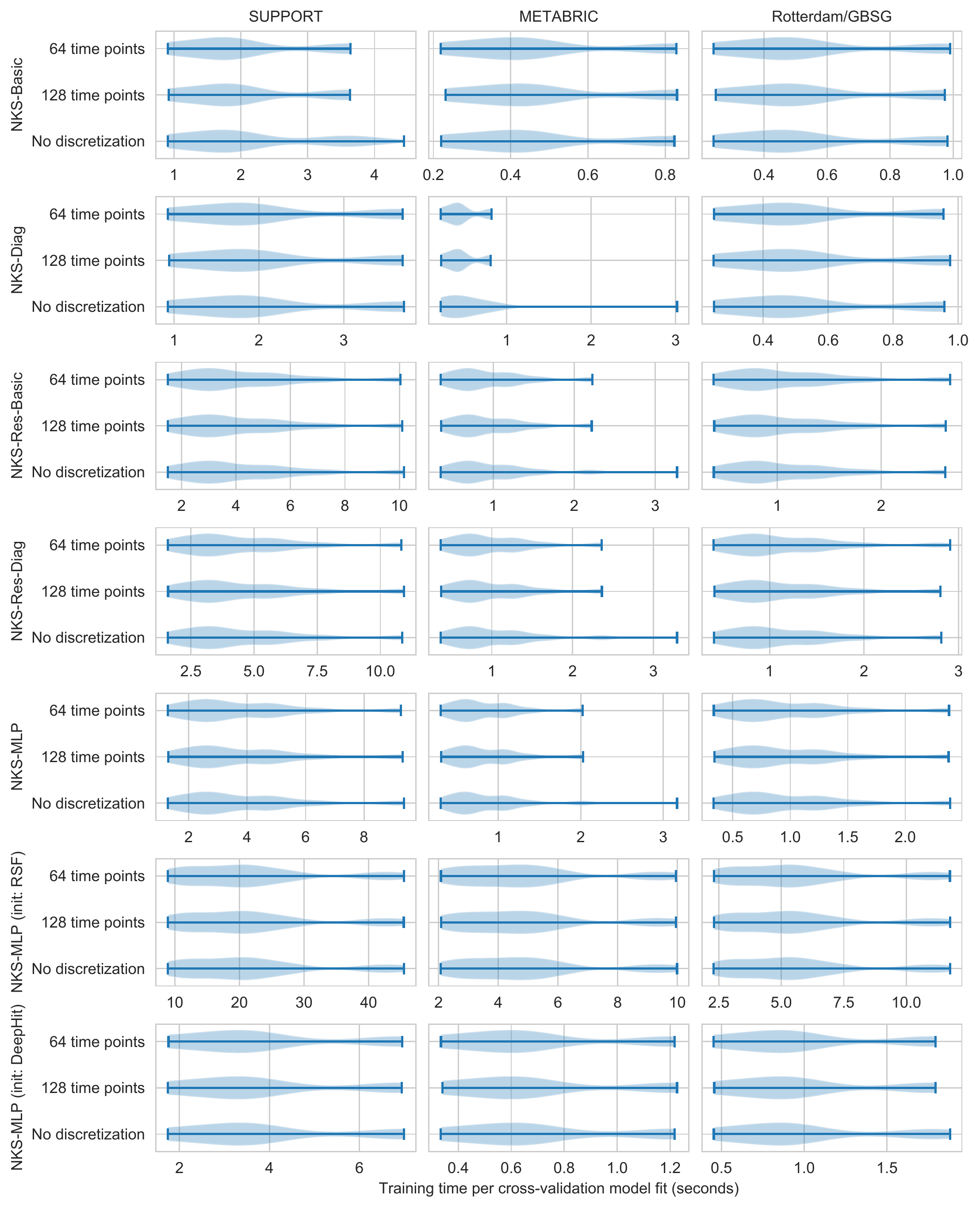}
\vspace{-2.5em}
\caption{Distributions of cross-validation model training times across datasets and \textsc{nks} variants. The times for \textsc{nks-mlp} with \textsc{rsf}/\textsc{deephit} initializations exclude training times of \textsc{rsf}/\textsc{deephit}.}
\label{fig:n-durations-cv-fitting-times}
\vspace{-1em}
\end{figure}

For the cross-validation model training times shown in Figure~\ref{fig:cv-fitting-times}, we further subdivide the training times of the \textsc{nks} variants depending on the time grid quantization level (no quantization vs using 64 or 128 time points) to obtain the distributions of cross-validation training times in Figure~\ref{fig:n-durations-cv-fitting-times}. We find that at least for the datasets we tested on, while quantizing to fewer time points occasionally reduces computation time, very often the difference in training times between the quantization levels is negligible. We suspect that for datasets with significantly larger numbers of unique observed times in the training data and where mini-batch training is used with large batch sizes, then the quantization level might have a more dramatic effect on training times.

\end{document}